\documentclass{article}

\usepackage[preprint]{neurips_2024}
\usepackage[utf8]{inputenc}
\usepackage[T1]{fontenc}
\usepackage{hyperref}
\usepackage{url}
\usepackage{booktabs}
\usepackage{amsmath}
\usepackage{amssymb}
\usepackage{graphicx}
\usepackage{xcolor}
\usepackage{enumitem}

\title{Prototype Latent World Model Replay for Class-Incremental Learning}

\author{
  Weizhi Nie, Hui Wang, Weijie Wang, and Yuting Su\\
  Tianjin University\\
  \texttt{weizhinie@tju.edu.cn}
}

\begin{document}

\maketitle

\begin{abstract}
Class-incremental learning requires a model to learn new classes while preserving decision regions for old ones.
This is difficult when raw old samples are no longer available.
We propose \emph{Prototype Latent World Model Replay}, a memory-free framework that stores old classes as distributions over stable hidden states rather than as images.
A frozen ImageNet-pretrained encoder maps each image into a latent state space.
In this space, each class is summarized by several prototype-centered distributions with class-specific variances.
When new classes arrive, the model samples old latent states from this prototype world model.
It then trains a lightweight adapter and classifier using both sampled old states and real new-class features.
We also add a supervised contrastive term in the adapter space to promote intra-class compactness and old-new class separation.
On Split CIFAR-100, our method improves over fine-tuning under Inc5, Inc10, and Inc20 without storing raw exemplars.
The full Ours-LWM+Con model raises LastAcc from 4.55\% to 31.64\%, from 9.06\% to 37.06\%, and from 16.96\% to 43.10\% in Inc5, Inc10, and Inc20, respectively.
It also achieves AvgAcc of 45.86\%, 52.19\%, and 56.18\%.
Ablation and retention analyses show that stable latent-state replay is the main source of the gain.
Contrastive separation further refines the old-new geometry.
These results suggest that prototype latent memory preserves reusable class-state distributions, rather than only fitting the current classifier.
\end{abstract}

\section{Introduction}

Class-incremental learning (CIL) aims to build visual recognition systems that can learn new categories over time.
The model should do this without retraining on all earlier data.
This setting is central to deployment.
A model may first learn one set of objects and later encounter new categories as the environment, user needs, or data stream changes.
Unlike standard supervised learning, CIL requires the model to classify all classes seen so far.
During training, however, only the current task data are available.
This absence of old-class samples makes neural networks vulnerable to catastrophic forgetting.
In this failure mode, learning new classes overwrites the decision regions of old classes.

The challenge is not only that the classifier changes.
The semantic space used by the classifier can also drift.
When a model is fine-tuned on new classes, old images are unavailable.
They therefore cannot anchor their previous feature regions.
As a result, new-class representations may move into old-class regions.
Classifier boundaries then become biased toward the latest task, and old classes may collapse to near-zero accuracy.
This old-new interference grows as more tasks arrive.
It is one of the central obstacles in class-incremental learning.

Existing methods address forgetting from several directions.
Regularization-based methods constrain parameter changes or distill predictions from a previous model.
Rehearsal-based methods store a limited number of old examples and interleave them with new data.
Generative replay methods attempt to synthesize old samples when raw data are unavailable.
These strategies have made important progress, but they also have limits.
Parameter regularization does not explicitly preserve old-class distributions.
Exemplar memory can be constrained by storage or privacy.
Image-level generation can be costly and unstable.
This motivates a simple question: can a model remember old classes without storing or regenerating old images?

We approach this question through latent state modeling.
Although raw images are unavailable in later tasks, old classes may still be represented by their semantic states.
If the latent space is stable across tasks, old-class knowledge can be stored as a compact distribution over latent states.
New learning can then be constrained by replaying sampled old states instead of images.
This view is inspired by world models.
Rather than memorizing individual observations, the learner maintains a structured model of possible states for each class.

\begin{figure}[t]
    \centering
    \includegraphics[width=\linewidth]{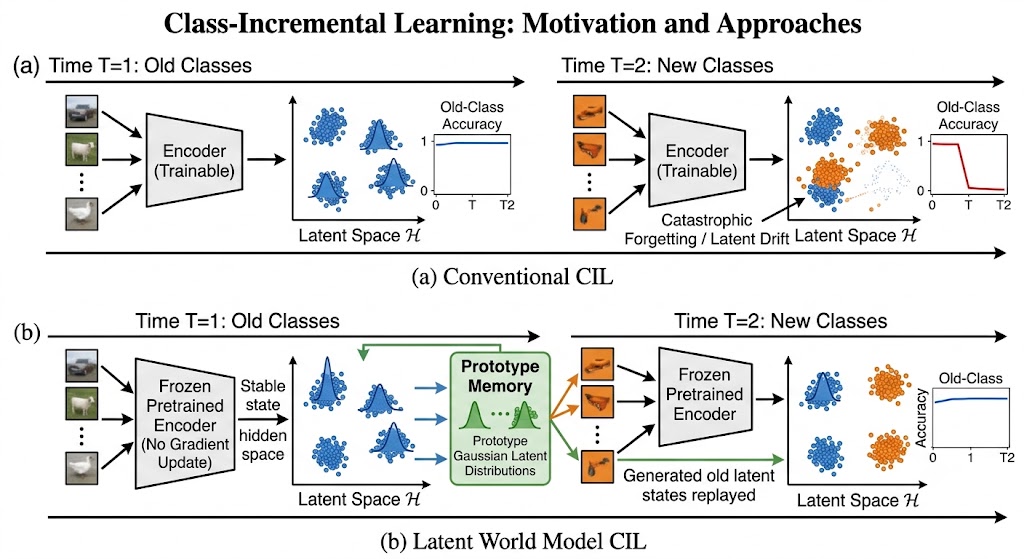}
    \caption{
    Motivation of the proposed framework.
    Conventional fine-tuning updates the representation and classifier using only new-class samples, causing old latent regions to drift or collapse.
    We instead maintain old classes as prototype-centered distributions in a stable latent state space $H$.
    During later tasks, sampled old latent states are replayed through a trainable adapter and classifier, while a class-separable constraint discourages new classes from invading old-class regions.
    }
    \label{fig:motivation}
\end{figure}

Based on this motivation, we propose \emph{Prototype Latent World Model Replay}.
The method first maps images into a stable latent state space $H$ with a frozen pretrained encoder.
In this space, each class is not represented by a single point.
Instead, it is represented by a mixture of prototype-centered distributions that capture multiple modes of intra-class variation.
When a new task arrives, the model samples latent states from old-class prototype distributions.
It replays these states together with real new-class examples.
The sampled states pass through a trainable adapter and classifier.
This allows task-specific adaptation while keeping the underlying latent world memory stable.

Two design principles follow from this formulation.
First, the latent state space should be stable enough that stored old-class distributions remain meaningful across tasks.
We therefore freeze the pretrained encoder and update only the adapter and classifier during incremental learning.
Second, old and new classes should remain separable in the adapter space.
We introduce a supervised contrastive term over real new-class features and generated old-class latent features.
It pulls same-class states together and pushes different-class states apart.
This constraint addresses old-new interference by discouraging new-class features from occupying old-class semantic regions.

In summary, our method treats forgetting as a problem of preserving old-class latent state distributions under new-class adaptation.
Rather than relying on raw old images, we store a compact prototype world memory in $H$ and use sampled latent states to constrain subsequent learning.
This formulation provides a simple bridge between pretrained representation learning, latent replay, and class-incremental recognition.

\paragraph{Contributions.}
We summarize our contributions as follows:
\begin{itemize}[leftmargin=1.2em]
    \item We propose a latent world model replay framework for class-incremental learning. It preserves old classes as distributions over stable pretrained latent states rather than as raw images.
    \item We introduce a prototype-based class memory. It represents each class with multiple latent prototypes and variances, and combines latent replay with supervised contrastive separation.
    \item We evaluate the method on Split CIFAR-100 with representative baselines, ablations, and interpretability analyses, including task-retention visualization and prototype coverage.
\end{itemize}

\section{Related Work}

\subsection{Catastrophic Forgetting and Class-Incremental Learning}

Continual learning studies how neural networks learn from a non-stationary sequence of tasks without losing old capabilities.
Catastrophic forgetting has been measured in many settings.
These studies show that standard neural networks can overwrite old knowledge when they are optimized only on new data \citep{kirkpatrick2017ewc,kemker2018measuring,parisi2019continual,wang2024survey}.
Class-incremental learning is one of the most challenging variants.
Task identity is unavailable at inference time, and the classifier must discriminate among all classes seen so far.
Early methods such as iCaRL combine exemplar memory, nearest-mean classification, and distillation to maintain old-class recognition \citep{rebuffi2017icarl}.
Later work improved classifier calibration, representation preservation, and memory efficiency.
However, the central difficulty remains: old classes are not represented by real data during later tasks, while new-class samples dominate optimization.

The forgetting problem has two coupled aspects.
First, feature representations may drift when the backbone is updated on new classes.
Old-class regions then no longer align with their learned decision boundaries.
Second, even when old features are partly preserved, the classifier can become biased toward recent classes.
Methods such as LUCIR-style rebalancing, bias correction, weight alignment, and pooled-output distillation address these effects with classifier or feature constraints \citep{hou2019lucir,wu2019bic,zhao2020wa,douillard2020podnet}.
Our work shares the goal of preserving old-class regions.
However, we move the memory object from raw images or logits to distributions over a stable latent state space.

\subsection{Regularization, Distillation, and Classifier Calibration}

Regularization-based continual learning methods limit changes to parameters that matter for previous tasks.
Elastic Weight Consolidation estimates parameter importance with a Fisher approximation \citep{kirkpatrick2017ewc}.
Synaptic Intelligence accumulates path-integral importance during training \citep{zenke2017si}.
Memory Aware Synapses preserve parameters that influence the network output \citep{aljundi2018mas}.
These methods are attractive because they do not store old examples.
However, parameter-level constraints are indirect for class-incremental recognition.
Preserving weights does not necessarily preserve old-class feature distributions or calibrated decision regions.

Knowledge distillation provides another route.
Learning without Forgetting distills outputs from the previous model while training on new classes \citep{li2018lwf}.
Later class-incremental methods distill intermediate features or spatial activation patterns to preserve representations \citep{douillard2020podnet}.
End-to-end incremental learning combines distillation with exemplar memory and balanced fine-tuning \citep{castro2018eeil}.
Bias-correction methods note that new-class logits are often overestimated.
They therefore correct the classifier head after incremental training \citep{wu2019bic,zhao2020wa}.
These approaches reduce forgetting, but they often rely on a previous teacher, an exemplar buffer, or post-hoc calibration.
In contrast, our method builds an explicit latent-state memory for old classes.
It uses sampled old states to regularize the adapter and classifier during new-task learning.

\subsection{Rehearsal, Generative Replay, and Latent Replay}

Rehearsal methods remain strong practical solutions for class-incremental learning.
They directly expose the model to old-class information.
Gradient Episodic Memory and A-GEM constrain gradients using stored samples \citep{lopezpaz2017gem,chaudhry2019agem}.
iCaRL and many later CIL methods maintain exemplar buffers to approximate the old data distribution \citep{rebuffi2017icarl,castro2018eeil,buzzega2020der}.
Dark Experience Replay shows that simple replay with stored logits is a strong continual learning baseline \citep{buzzega2020der}.
However, raw-example rehearsal can be limited by memory budgets, privacy constraints, or deployment settings where old data cannot be retained.

Generative replay avoids storing old images by training a generator to synthesize previous-task data \citep{shin2017dgr,vandeven2018feedback,vandeven2020brain}.
This idea is close to our goal of replaying unavailable old knowledge.
However, image-level generative replay must model high-dimensional pixel distributions.
It can be costly, and artifacts may harm classifier training.
Feature-level and exemplar-free methods offer a lighter alternative by manipulating representations instead of images.
FeTrIL translates features to approximate unseen old-class examples \citep{petit2023fetril}.
PASS uses prototype augmentation and self-supervision for exemplar-free learning \citep{zhu2021pass}.
SSRE expands representations without exemplar storage \citep{zhu2022ssre}.
Generative classifiers model class-conditional feature distributions for incremental recognition \citep{van2021generativeclassifiers}.
Our method belongs to this broader feature-replay family, but it differs in two ways.
It uses a frozen pretrained encoder to define a stable state space.
It also models each class as a multi-prototype latent world distribution rather than as a single feature template.
Recent work has explored multimodal and graph-structured incremental or open-category learning.
Examples include causal intervention for multimodal class-incremental 3D recognition, incremental graph fusion for anomaly detection, and cross-domain discovery of new fault categories \citep{li2025causalcil3d,ma2026mfgcn,wang2024newfault}.
These studies show the value of structured state information when new categories or domains appear.
Our work stores such information as replayable distributions in a stable visual latent space.

\subsection{Pretrained Representations, Prototypes, and Contrastive Objectives}

Pretrained visual representations have changed class-incremental learning.
Strong pretrained backbones provide transferable features and reduce the need for full backbone updates.
Several methods use frozen or lightly adapted pretrained models for CIL \citep{wu2022strongptm,zhou2023simplecil,mcdonnell2023ranpac,zhou2024ease}.
FOSTER and MEMO also highlight representation expansion, compression, and memory-efficient adaptation in incremental recognition \citep{wang2022foster,zhou2023memo}.
These studies suggest that a stable representation space can reduce forgetting.
They do not, however, explicitly model old classes as latent state distributions.
We build on this observation by treating the pretrained encoder output as a stable $H$-space.
Old classes can then be stored as prototype mixtures in that space.

Prototype-based representations are natural for CIL because they summarize class structure compactly.
Instead of storing all old samples, one can preserve class centers, translated features, or prototype distributions.
Our method extends this principle by using multiple prototypes per class.
Each prototype has an estimated variance, which helps capture intra-class modes in the latent state space.
This design is closer to a latent world model than to a nearest-class-mean classifier.
The memory can sample plausible old-class states during later tasks.

Self-supervised and contrastive learning provide tools for shaping representation geometry.
Contrastive methods such as SimCLR and MoCo learn representations by pulling related views together and pushing unrelated samples apart \citep{chen2020simclr,he2020moco}.
BYOL, DINO, MAE, I-JEPA, and DINOv2 show that predictive or self-distilled objectives can produce strong visual features without labels \citep{grill2020byol,caron2021dino,he2022mae,assran2023ijepa,oquab2023dinov2}.
Supervised contrastive learning extends this idea by using labels to compact same-class features and separate different classes \citep{khosla2020supcon}.
We adopt this principle in the adapter space over both real new-class features and generated old-class latent features.
The resulting loss targets old-new interference directly.
Generated old states and real new states should not collapse into the same decision region.

\section{Our Approach}

\subsection{Overview}

\begin{figure}[t]
    \centering
    \includegraphics[width=\linewidth]{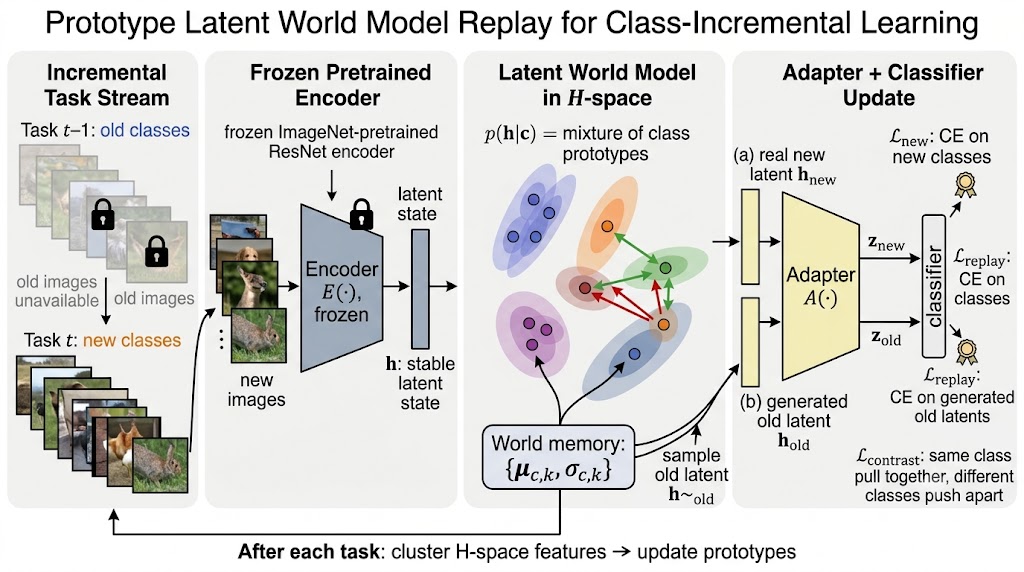}
    \caption{
    Framework of Prototype Latent World Model Replay.
    A frozen pretrained encoder maps images to stable latent states $h \in H$.
    After each task, class-wise latent states are clustered into multiple prototype distributions and stored as a latent world memory.
    During later tasks, old-class latent states are sampled from this memory and replayed together with real new-class samples through a trainable adapter and classifier.
    A supervised contrastive objective further encourages same-class compactness and different-class separation in the adapter space.
    }
    \label{fig:framework}
\end{figure}

Figure~\ref{fig:framework} illustrates the proposed framework.
Our goal is to learn new classes while preserving old-class decision regions without storing raw old images.
The key idea is to separate the learning system into two spaces.
The first is a stable latent state space $H$ produced by a frozen pretrained encoder.
This space stores old-class knowledge as a compact latent world memory.
The second is a task-adaptive classification space $Z$ produced by a lightweight trainable adapter.
The adapter and classifier can adapt to new tasks.
At the same time, the old-class memory in $H$ remains stable.

The design is motivated by two observations.
First, catastrophic forgetting is closely related to representation drift.
If old-class regions move or collapse when new classes arrive, the previous classifier boundaries become invalid.
Second, old raw images are often unavailable in CIL.
A stable pretrained representation can still provide a coordinate system in which old classes can be described statistically.
We therefore store old-class distributions in $H$ instead of storing images.
During later tasks, we sample old latent states from this memory.

\subsection{Problem Formulation and Notation}

We consider a class-incremental learning scenario with a sequence of $T$ tasks:
\begin{equation}
    \mathcal{T}_1, \mathcal{T}_2, \ldots, \mathcal{T}_T .
\end{equation}
Each task $\mathcal{T}_t$ provides a training set
\begin{equation}
    \mathcal{D}_t = \{(x_i, y_i)\}_{i=1}^{N_t},
\end{equation}
where $x_i$ is an input image, $y_i$ is its class label, and $N_t$ is the number of current-task samples.
The class set of task $t$ is denoted by $\mathcal{C}_t$.
Following the standard CIL protocol, class sets are disjoint:
\begin{equation}
    \mathcal{C}_i \cap \mathcal{C}_j = \emptyset, \quad i \neq j .
\end{equation}
At training stage $t$, the model can access only the current dataset $\mathcal{D}_t$.
Old datasets $\mathcal{D}_1,\ldots,\mathcal{D}_{t-1}$ are unavailable.
At test time, the model must classify among all observed classes:
\begin{equation}
    \mathcal{C}_{1:t} = \bigcup_{j=1}^{t} \mathcal{C}_j .
\end{equation}

We use $E(\cdot)$ to denote the pretrained encoder, $A(\cdot)$ to denote the trainable adapter, and $G(\cdot)$ to denote the classifier.
The encoder maps an image to a latent state:
\begin{equation}
    h = E(x), \quad h \in H,
\end{equation}
where $H$ is the stable latent state space.
The adapter maps $h$ to a task-adaptive representation:
\begin{equation}
    z = A(h), \quad z \in Z,
\end{equation}
where $Z$ is the classification feature space.
The classifier produces logits:
\begin{equation}
    o = G(z).
\end{equation}
Here $o$ is a vector of class scores over all classes observed so far.

\subsection{Stable Latent State Space}

The first component of our method is the stable latent state space $H$.
We instantiate $E(\cdot)$ as an ImageNet-pretrained encoder and freeze its parameters during incremental training.
Freezing the encoder is important because stored old-class distributions are meaningful only if the coordinate system of $H$ stays fixed.
If the encoder were updated freely on new classes, a stored old prototype could point to a region with a different semantic meaning.
We also observed this mismatch in preliminary experiments.
Latent replay becomes ineffective when the encoder representation drifts.

The adapter $A(\cdot)$ prevents the system from becoming too rigid.
Although $H$ is fixed, the classifier must still adapt to a changing class set.
The adapter provides a small trainable transformation from $H$ to the classification space $Z$.
Thus, $E$ preserves semantic coordinates, while $A$ and $G$ provide plasticity for new tasks.

\subsection{Prototype Latent World Memory}

After training on task $t$, we build a memory for each class $c \in \mathcal{C}_t$.
Let
\begin{equation}
    \mathcal{H}_c = \{h_i = E(x_i) \mid y_i = c\}
\end{equation}
be the set of latent states for class $c$.
Rather than representing a class by a single center, we cluster $\mathcal{H}_c$ into $K$ groups.
The motivation is that a visual class can contain several modes, such as different poses, colors, or backgrounds.
A single prototype may blur these modes.
Multiple prototypes better approximate the class manifold.

For the $k$-th cluster of class $c$, we compute its mean and diagonal variance:
\begin{equation}
    \mu_{c,k} = \frac{1}{|\mathcal{I}_{c,k}|}
    \sum_{i \in \mathcal{I}_{c,k}} h_i ,
\end{equation}
\begin{equation}
    \sigma_{c,k}^{2} =
    \frac{1}{|\mathcal{I}_{c,k}|}
    \sum_{i \in \mathcal{I}_{c,k}}
    (h_i - \mu_{c,k}) \odot (h_i - \mu_{c,k}),
\end{equation}
where $\mathcal{I}_{c,k}$ is the index set of samples assigned to prototype $k$, and $\odot$ denotes element-wise multiplication.
The memory item for this prototype is
\begin{equation}
    \mathcal{M}_{c,k} = (\mu_{c,k}, \sigma_{c,k}^{2}).
\end{equation}
The complete memory of class $c$ is
\begin{equation}
    \mathcal{M}_{c} =
    \{(\mu_{c,k}, \sigma_{c,k}^{2})\}_{k=1}^{K}.
\end{equation}

This memory defines a class-conditional latent distribution:
\begin{equation}
    p(h \mid c)
    =
    \frac{1}{K}
    \sum_{k=1}^{K}
    \mathcal{N}
    \left(
    h;
    \mu_{c,k},
    \mathrm{diag}(\sigma_{c,k}^{2})
    \right).
\end{equation}
We call this distribution a prototype latent world model because it describes possible latent states of class $c$ in the stable space $H$.
It does not reconstruct pixels.
Instead, it models the semantic state distribution needed for recognition.

\subsection{Latent World Model Replay}

At task $t>1$, old images are unavailable.
For each old class $c \in \mathcal{C}_{1:t-1}$, we sample synthetic old latent states from the stored world memory:
\begin{equation}
    \tilde{h}_{old} \sim p(h \mid c).
\end{equation}
The sampled state $\tilde{h}_{old}$ is not an image and does not pass through the encoder.
It is already a point in $H$.
We therefore feed it directly into the adapter and classifier:
\begin{equation}
    \tilde{z}_{old} = A(\tilde{h}_{old}),
    \quad
    \tilde{o}_{old} = G(\tilde{z}_{old}).
\end{equation}
The replay loss is
\begin{equation}
    \mathcal{L}_{replay}
    =
    \mathrm{CE}(\tilde{o}_{old}, y_{old}),
\end{equation}
where $y_{old}$ is the label of the sampled old class.

This replay mechanism serves two purposes.
First, it gives the classifier explicit old-class training signals even when old images are absent.
Second, replay happens in the stable state space $H$.
It therefore avoids the difficulty of generating high-dimensional images.
In practice, we sample old latent states in a class-balanced manner.
This ensures that all previous classes contribute to each training stage.

\subsection{New-Class Learning}

For current-task images $(x_{new}, y_{new}) \in \mathcal{D}_t$, the model performs ordinary supervised learning.
The forward path is
\begin{equation}
    h_{new} = E(x_{new}),
    \quad
    z_{new} = A(h_{new}),
    \quad
    o_{new} = G(z_{new}).
\end{equation}
The new-class classification loss is
\begin{equation}
    \mathcal{L}_{new}
    =
    \mathrm{CE}(o_{new}, y_{new}).
\end{equation}
This term provides plasticity.
It teaches the adapter and classifier to recognize newly arriving classes.
However, using $\mathcal{L}_{new}$ alone would bias the model toward the latest task.
This is why latent replay is needed.

\subsection{Class-Separable Contrastive Constraint}

Latent replay preserves old-class supervision.
However, replay alone does not explicitly enforce geometric separation between old and new classes.
New-class features may still move close to old-class states in the adapter space $Z$.
To reduce this interference, we add a supervised contrastive constraint.

For each mini-batch, we combine real new-class features and generated old-class features:
\begin{equation}
    \mathcal{B}
    =
    \{(z_i, y_i)\}_{i=1}^{B},
    \quad
    z_i \in \{z_{new}\} \cup \{\tilde{z}_{old}\}.
\end{equation}
For an anchor feature $z_i$, the positive set is
\begin{equation}
    P(i) = \{p \neq i \mid y_p = y_i\}.
\end{equation}
Using cosine similarity $\mathrm{sim}(\cdot,\cdot)$ and temperature $\tau$, the supervised contrastive loss is
\begin{equation}
\mathcal{L}_{con}
=
- \sum_{i}
\frac{1}{|P(i)|}
\sum_{p \in P(i)}
\log
\frac{
\exp(\mathrm{sim}(z_i,z_p)/\tau)
}{
\sum_{a \neq i}
\exp(\mathrm{sim}(z_i,z_a)/\tau)
}.
\end{equation}
This term pulls features with the same label together and pushes features with different labels apart.
It is especially useful for old-new separation.
Generated old states and real new states are trained jointly in $Z$.
This reduces the chance that new classes overwrite old regions.

\subsection{Overall Objective and Training Procedure}

The overall objective for task $t$ is
\begin{equation}
    \mathcal{L}
    =
    \mathcal{L}_{new}
    +
    \lambda_{r}\mathcal{L}_{replay}
    +
    \lambda_{c}\mathcal{L}_{con},
\end{equation}
where $\lambda_r$ controls old latent replay and $\lambda_c$ controls class-separable contrastive regularization.
At task $1$, no old memory is available.
The model is therefore trained only with $\mathcal{L}_{new}$.
It then initializes the memory for the first set of classes.
For later tasks, each mini-batch contains real new-class samples and generated old latent states.

The training procedure is summarized as follows.
First, freeze the pretrained encoder $E$ to define the stable latent state space $H$.
Second, train the adapter $A$ and classifier $G$ on the current task.
This training uses new-class supervision, old-class latent replay, and contrastive separation.
Third, after the task is finished, extract $H$-space features for the current classes.
Then cluster them into $K$ prototypes per class, estimate prototype variances, and add them to the world memory.
This cycle repeats for all incremental tasks.

\section{Experiments}

\subsection{Datasets and Incremental Protocols}

We evaluate the proposed method on Split CIFAR-100, a standard class-incremental image classification benchmark.
Following common CIL practice \citep{masana2023cil,wang2024survey}, the dataset is split into a sequence of class-disjoint tasks.
After each stage, the model is evaluated on all classes observed so far.
Task identity is not provided at test time.

\paragraph{Split CIFAR-100.}
CIFAR-100 contains natural images of size $32 \times 32$ \citep{krizhevsky2009cifar}.
CIFAR-100 contains 100 classes with 50,000 training images and 10,000 test images, i.e., 500 training and 100 test images per class.
We consider Inc5, Inc10, and Inc20 protocols, where each incremental stage introduces 5, 10, or 20 new classes.
These settings create long, medium, and short incremental sequences.
They allow us to test severe forgetting under many small updates and adaptation under larger class batches.

For pretrained encoders, images are resized to the input resolution required by the backbone.
They are also normalized with the corresponding preprocessing statistics.
Unless otherwise specified, we use an ImageNet-pretrained ResNet-18 encoder \citep{he2016resnet}.
The encoder is frozen, and only the adapter and classifier are updated during incremental learning.

\subsection{Baselines and Compared Methods}

The first experiment compares our method with representative state-of-the-art CIL methods.
Following recent CIL papers, we report results under multiple incremental protocols.
We compare both average-stage performance and final-stage retention.
To clarify the comparison, we organize baselines into two groups.

\paragraph{Conventional CIL baselines.}
This group includes methods commonly evaluated with CNN backbones and standard CIL protocols.
We include fine-tuning, LwF \citep{li2018lwf}, iCaRL \citep{rebuffi2017icarl}, BiC \citep{wu2019bic}, LUCIR \citep{hou2019lucir}, PODNet \citep{douillard2020podnet}, DER \citep{buzzega2020der}, FOSTER \citep{wang2022foster}, FeTrIL \citep{petit2023fetril}, PASS \citep{zhu2021pass}, and SSRE \citep{zhu2022ssre}.
These methods cover distillation, exemplar rehearsal, classifier calibration, feature replay, and exemplar-free prototype-based learning.

\paragraph{Pretrained-model CIL baselines.}
Because our method uses a frozen pretrained representation, we also compare with methods that use pretrained models or prompt/adaptation mechanisms.
This group includes L2P \citep{wang2022l2p}, DualPrompt \citep{wang2022dualprompt}, CODA-Prompt \citep{smith2023coda}, SLCA \citep{zhang2023slca}, RanPAC \citep{mcdonnell2023ranpac}, SimpleCIL \citep{zhou2023simplecil}, and EASE \citep{zhou2024ease}.
These baselines are particularly relevant because they also study how pretrained representations improve continual learning.

\paragraph{Our variants.}
We report at least three variants of our method:
\begin{itemize}[leftmargin=1.2em]
    \item \textbf{Ours w/o replay}: pretrained encoder, adapter, and classifier trained only on current classes.
    \item \textbf{Ours-LWM}: latent world model replay with prototype memory but without the contrastive constraint.
    \item \textbf{Ours-LWM+Con}: the full method with prototype latent replay and supervised contrastive separation.
\end{itemize}
This organization lets the main table show two effects.
It reports the improvement over external baselines and the additional effect of the contrastive constraint.

\subsection{Evaluation Metrics}

Let $A_t$ denote the overall classification accuracy after task $t$ over all classes observed so far.
Following prior CIL work, we report Average Accuracy and Last Accuracy:
\begin{equation}
    \mathrm{AvgAcc}
    =
    \frac{1}{T}
    \sum_{t=1}^{T} A_t,
    \quad
    \mathrm{LastAcc}
    =
    A_T .
\end{equation}
Average Accuracy measures the learning trajectory across all stages, while Last Accuracy measures long-term retention after all tasks.
We also report average forgetting when per-task accuracies are available.
Let $a_{t,j}$ be the accuracy on task $j$ after learning task $t$.
The forgetting of task $j$ after the final task is
\begin{equation}
    F_j
    =
    \max_{t \in \{j,\ldots,T-1\}} a_{t,j}
    -
    a_{T,j}.
\end{equation}
The average forgetting is computed over all non-final tasks.
In addition, we report per-task accuracy matrices to diagnose which old tasks are preserved or forgotten.

\subsection{Comparison with State-of-the-Art Methods}

The first experiment evaluates whether latent world memory improves class-incremental recognition.
Inspired by recent CIL evaluation protocols, we report results on Split CIFAR-100 under Inc5, Inc10, and Inc20.
For each protocol, we report AvgAcc and LastAcc.

Table~\ref{tab:sota_cifar100} reports the reproduced CIFAR-100 comparison.
For a fair comparison, all methods use the same class order and train/test split.
Pretrained-model methods use the same pretrained backbone whenever possible.
For methods that require an exemplar buffer, we report the buffer size explicitly.
Because our main method does not store raw old images, its primary comparison is the memory-free setting.
We use a fast reproduction protocol with 10 epochs per incremental stage for all reproduced baselines.
This keeps the training budget comparable to our validation runs.
Unless otherwise noted, PyCIL baselines are trained from scratch with ResNet32-style CIFAR backbones.
Our method uses a frozen ImageNet-pretrained ResNet-18 encoder.

\begin{table}[t]
\centering
\caption{
Current comparison with state-of-the-art methods on Split CIFAR-100.
Inc5, Inc10, and Inc20 indicate 5, 10, and 20 new classes per incremental stage.
Avg and Last denote Average Accuracy and Last Accuracy.
Buffer denotes the total number of stored raw exemplars.
All reproduced baselines use the same class order and a 10-epoch-per-stage fast reproduction budget.
}
\label{tab:sota_cifar100}
\resizebox{\linewidth}{!}{
\begin{tabular}{lcccccccc}
\toprule
Method & Backbone & Buffer & \multicolumn{2}{c}{Inc5} & \multicolumn{2}{c}{Inc10} & \multicolumn{2}{c}{Inc20} \\
\cmidrule(lr){4-5}\cmidrule(lr){6-7}\cmidrule(lr){8-9}
 & & & Avg & Last & Avg & Last & Avg & Last \\
\midrule
Fine-tuning & Pretrained ResNet-18 & 0 & 16.37 & 4.55 & 25.00 & 9.06 & 37.87 & 16.96 \\
LwF~\citep{li2018lwf} & ResNet32 & 0 & 19.33 & 9.76 & 25.95 & 15.23 & 37.00 & 26.58 \\
iCaRL~\citep{rebuffi2017icarl} & ResNet32 & 2000 & 40.05 & 28.72 & 40.28 & 31.67 & 42.17 & 33.10 \\
BiC~\citep{wu2019bic} & ResNet32 & 2000 & 28.02 & 22.48 & 34.63 & 28.40 & 39.30 & 36.17 \\
PODNet~\citep{douillard2020podnet} & Cosine ResNet32 & 2000 & 19.69 & 9.98 & 32.79 & 22.45 & 36.67 & 27.36 \\
DER~\citep{buzzega2020der} & ResNet32 & 2000 & 40.43 & 32.75 & 45.60 & 37.47 & 46.27 & 43.62 \\
FeTrIL~\citep{petit2023fetril} & ResNet32 & 0 & 10.56 & 4.42 & 12.18 & 7.38 & 25.93 & 18.51 \\
SimpleCIL~\citep{zhou2023simplecil} & Cosine ResNet32 & 0 & 11.67 & 4.52 & 11.88 & 5.20 & 15.19 & 8.73 \\
\midrule
Ours-LWM & Pretrained ResNet-18 & 0 & 45.90 & 31.65 & 52.13 & 36.94 & 56.19 & 43.16 \\
Ours-LWM+Con & Pretrained ResNet-18 & 0 & 45.86 & 31.64 & 52.19 & 37.06 & 56.18 & 43.10 \\
\bottomrule
\end{tabular}
}
\end{table}

\begin{figure}[t]
    \centering
    \begin{minipage}{0.32\linewidth}
        \centering
        \includegraphics[width=\linewidth]{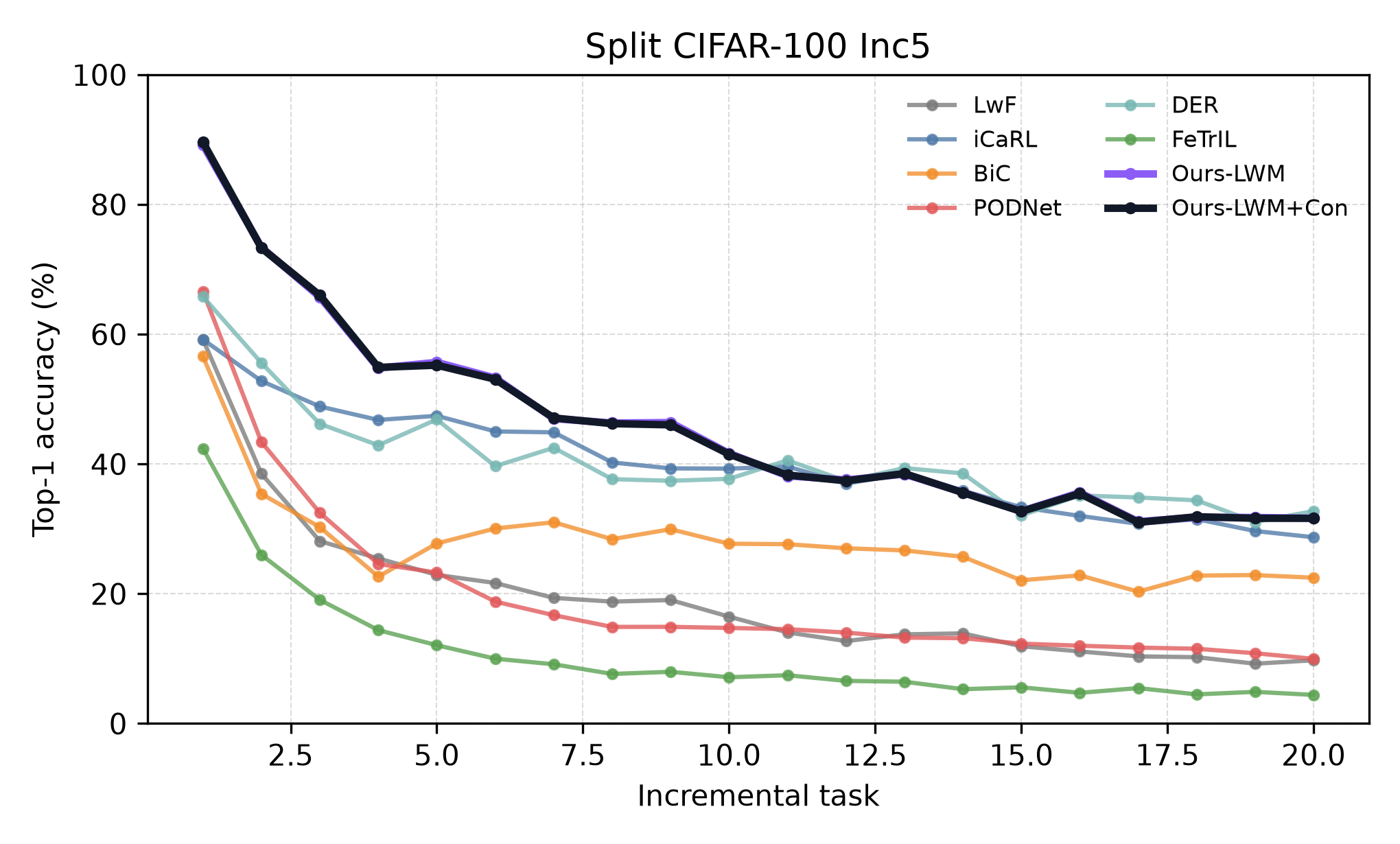}
    \end{minipage}
    \hfill
    \begin{minipage}{0.32\linewidth}
        \centering
        \includegraphics[width=\linewidth]{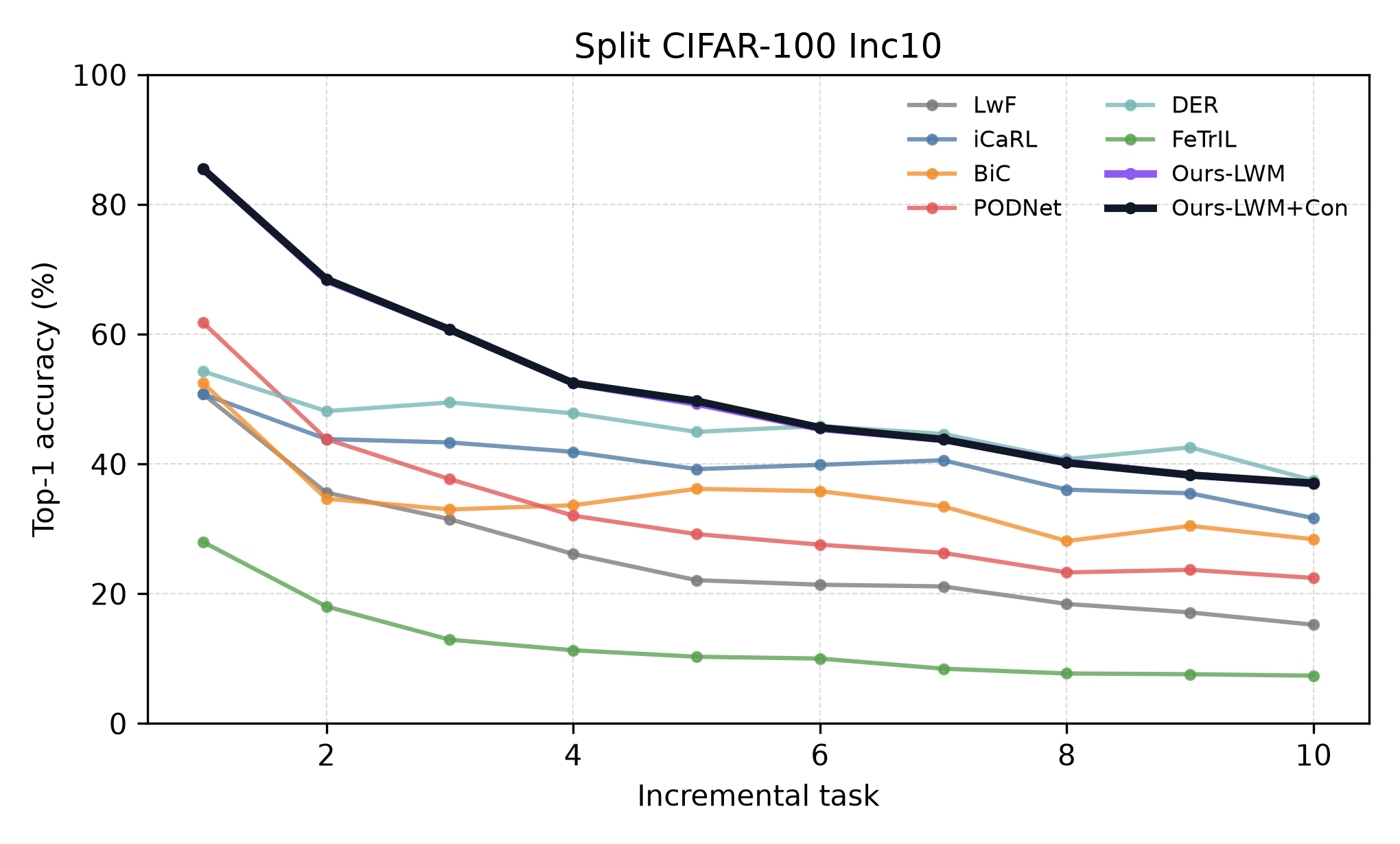}
    \end{minipage}
    \hfill
    \begin{minipage}{0.32\linewidth}
        \centering
        \includegraphics[width=\linewidth]{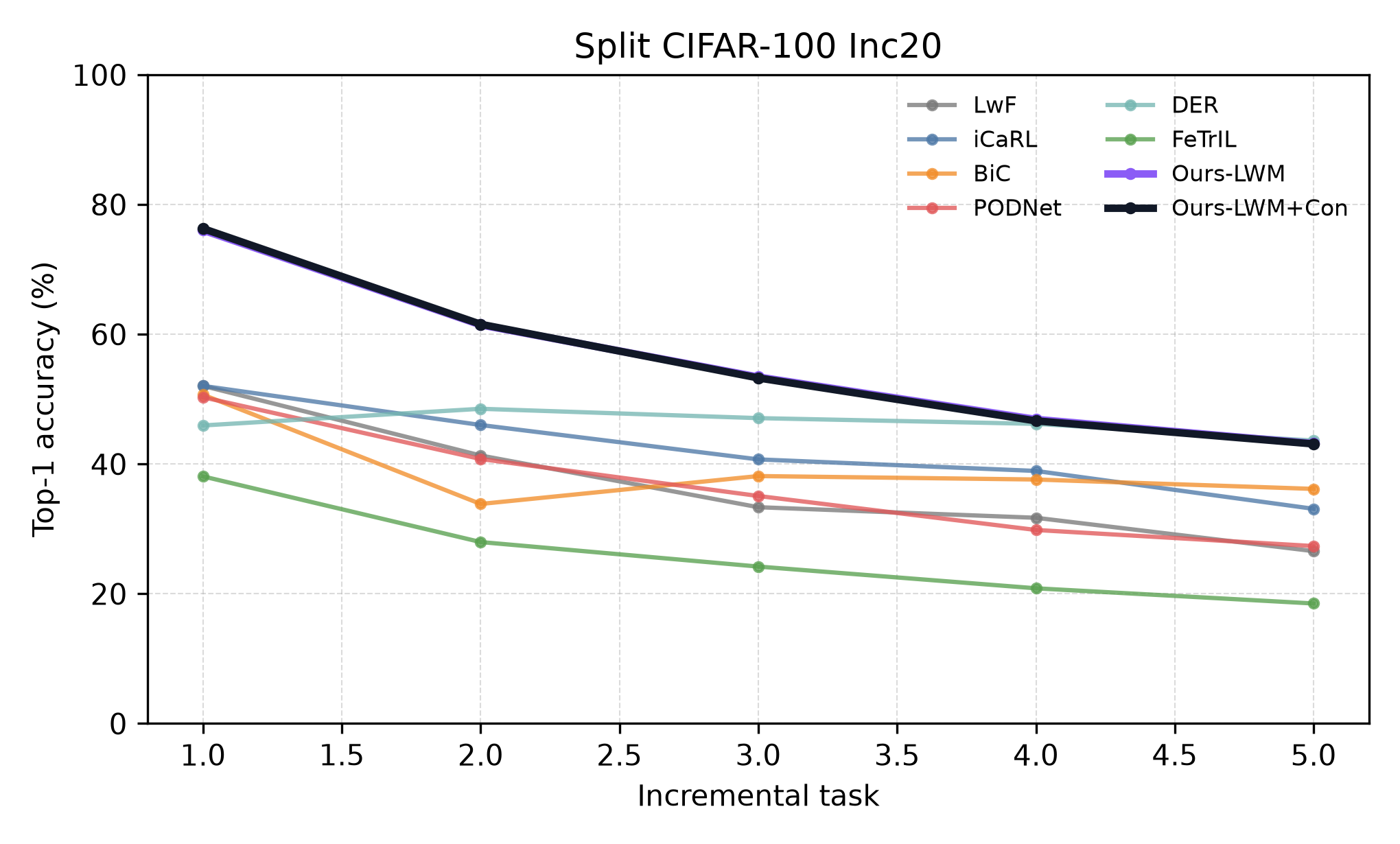}
    \end{minipage}
    \caption{
    Incremental top-1 accuracy curves on Split CIFAR-100.
    Each point reports the accuracy over all classes seen so far after one incremental training stage.
    Our latent world memory methods maintain substantially higher accuracy than fine-tuning and memory-free distillation/prototype baselines, while remaining competitive with strong exemplar-based methods.
    }
    \label{fig:cifar100_curves}
\end{figure}

Figure~\ref{fig:cifar100_curves} plots the full incremental accuracy curves.
These curves are useful because a method may perform well early but still forget after many increments.
Latent world memory slows the degradation curve, especially in Inc5 and Inc10.
These protocols contain more updates.
Among exemplar-based baselines, DER is the strongest competitor.
However, DER uses a 2000-image raw exemplar buffer and dynamically expands the representation.
In contrast, Ours-LWM and Ours-LWM+Con store only latent prototype statistics and no raw old images.
They still match or exceed DER in AvgAcc and remain competitive in LastAcc.

The key quantities are not only AvgAcc and LastAcc.
We also examine how accuracy degrades as the number of seen classes grows.
If the method preserves a latent world model, it should decline more slowly than fine-tuning.
It should also retain old tasks more clearly in the per-task accuracy matrix.
The CIFAR-100 results support this trend across all three protocols.
For example, in Inc5, fine-tuning reaches only 4.55\% LastAcc because earlier tasks collapse to nearly zero accuracy.
Ours-LWM+Con reaches 31.64\% LastAcc without storing raw old images.
In Inc10 and Inc20, it improves LastAcc from 9.06\% to 37.06\% and from 16.96\% to 43.10\%, respectively.
Compared with memory-free baselines such as LwF, FeTrIL, and SimpleCIL, our method obtains higher final accuracy under the fast protocol.
This suggests that replayable latent-state distributions are more effective than output distillation or static feature prototypes alone.
Compared with exemplar-based baselines such as iCaRL, BiC, PODNet, and DER, our method removes raw-image storage while preserving a competitive old-class decision space.

\subsection{Ablation Studies}

We next analyze which parts of the framework reduce forgetting.
The central question is whether the model needs a stable hidden state space.
We also test whether the gain can be obtained by simply fine-tuning a classifier on new classes.
We therefore conduct a full CIFAR-100 ablation under Inc5, Inc10, and Inc20.
All variants use the same class order, ImageNet-pretrained ResNet-18 feature extractor, 10 epochs per stage, and adapter/classifier capacity.
They differ in whether old classes are represented in the stable latent state space $H$.
They also differ in whether supervised contrastive separation shapes the adapter space.

\begin{table}[t]
\centering
\caption{
Full CIFAR-100 ablation under Inc5, Inc10, and Inc20.
Avg and Last denote Average Accuracy over incremental stages and final accuracy after all stages.
``No Stable $H$ Memory'' corresponds to ordinary fine-tuning without latent world memory.
}
\label{tab:ablation_components}
\resizebox{\linewidth}{!}{
\begin{tabular}{lccccccp{5.2cm}}
\toprule
Variant & \multicolumn{2}{c}{Inc5} & \multicolumn{2}{c}{Inc10} & \multicolumn{2}{c}{Inc20} & Interpretation \\
\cmidrule(lr){2-3}\cmidrule(lr){4-5}\cmidrule(lr){6-7}
 & Avg & Last & Avg & Last & Avg & Last & \\
\midrule
No Stable $H$ Memory & 16.37 & 4.55 & 25.00 & 9.06 & 37.87 & 16.96 & No old latent-state memory; the classifier is dominated by the newest classes. \\
Stable $H$ + Replay & 45.90 & 31.65 & 52.13 & 36.94 & 56.19 & 43.16 & Old classes are preserved as prototype latent distributions and replayed during later stages. \\
Stable $H$ + Replay + Con & 45.86 & 31.64 & 52.19 & 37.06 & 56.18 & 43.10 & Supervised contrastive separation shapes the old-new decision geometry. \\
\bottomrule
\end{tabular}
}
\end{table}

\begin{figure}[t]
    \centering
    \begin{minipage}{0.47\linewidth}
        \centering
        \includegraphics[width=\linewidth]{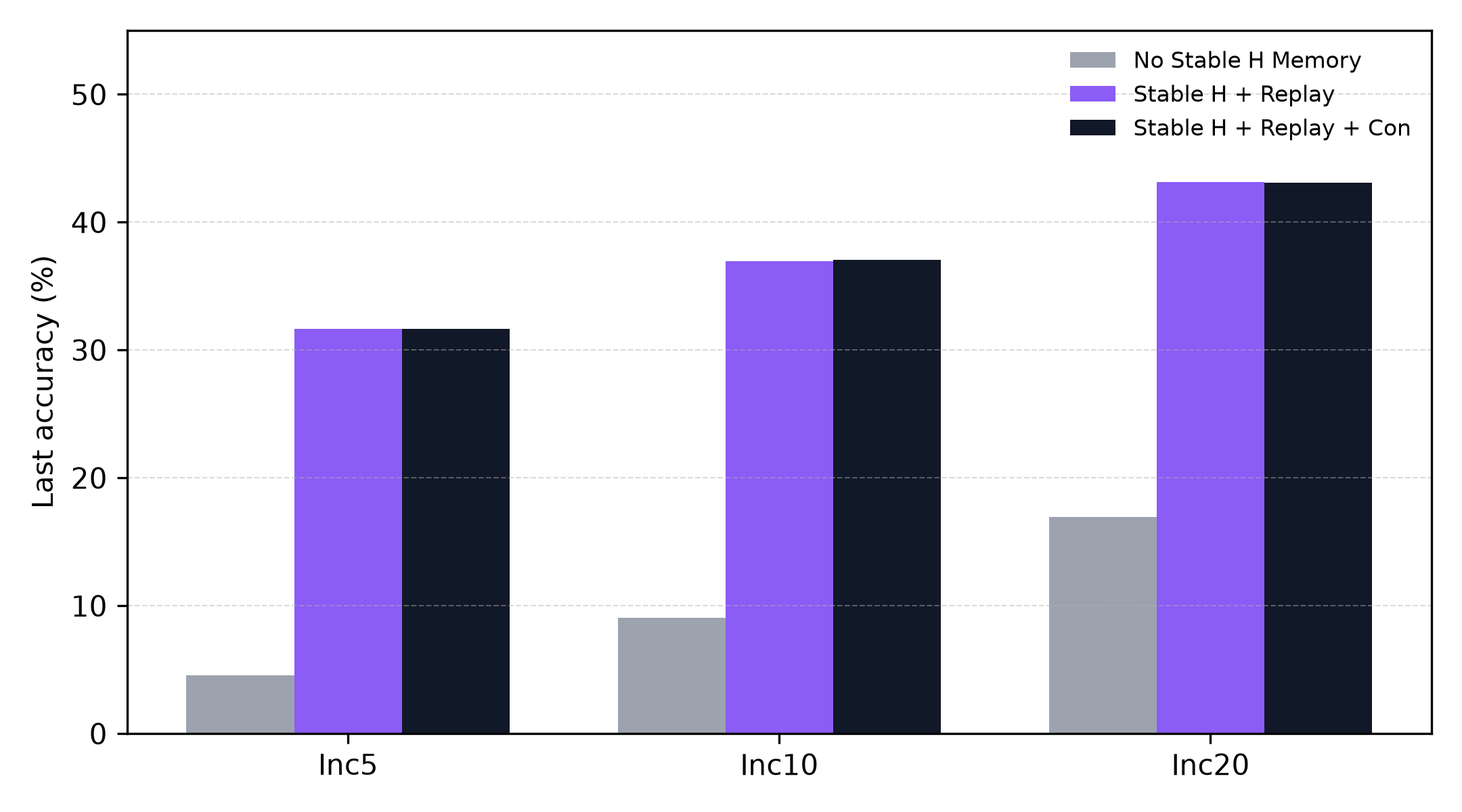}
        \vspace{-0.5em}
        {\small (a) Final accuracy across protocols}
    \end{minipage}
    \hfill
    \begin{minipage}{0.49\linewidth}
        \centering
        \includegraphics[width=\linewidth]{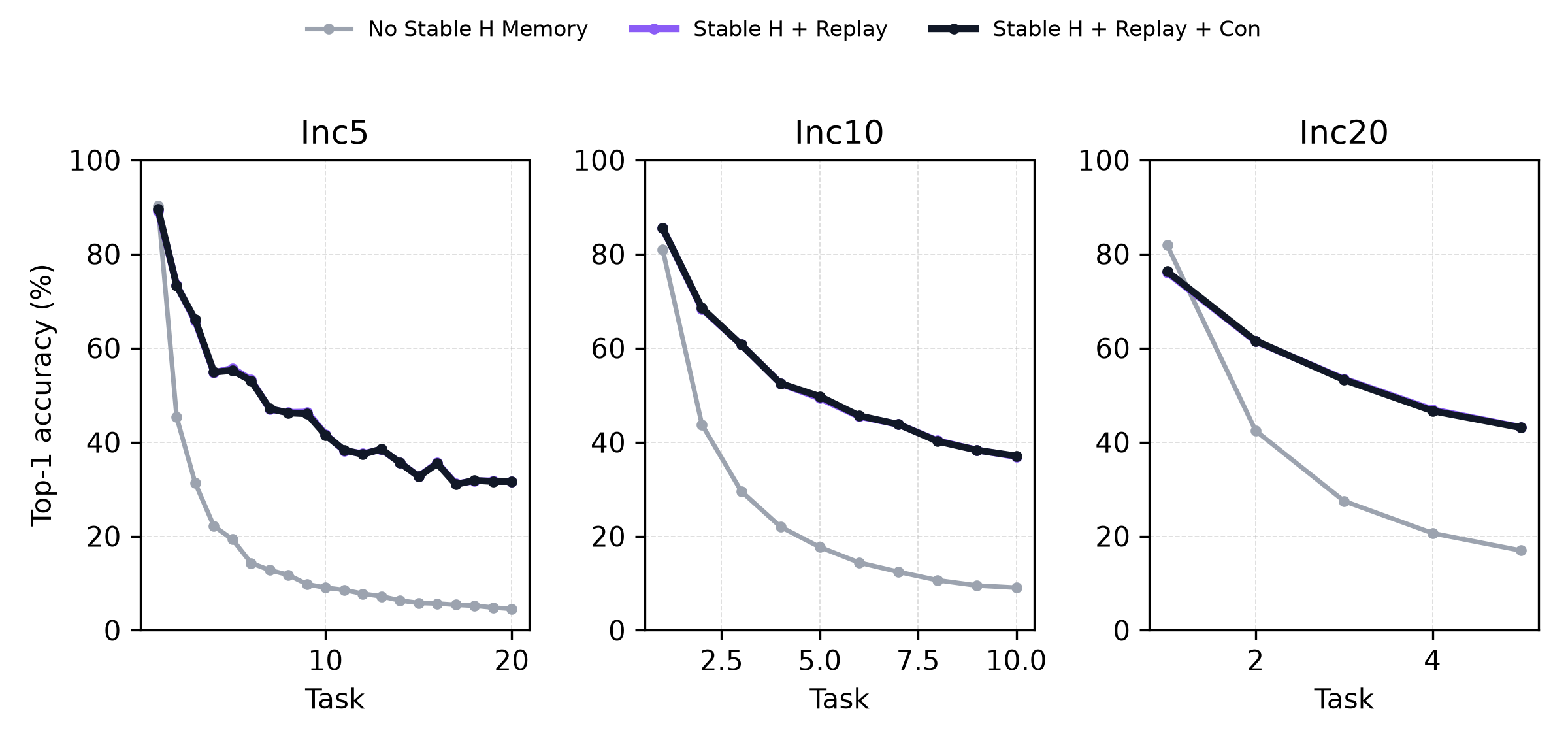}
        \vspace{-0.5em}
        {\small (b) Full incremental curves}
    \end{minipage}
    \caption{
    Full CIFAR-100 ablation analysis of stable latent world memory and supervised contrastive separation.
    Stable $H$-space replay gives the dominant gain by preventing old-class decision regions from disappearing.
    The supervised contrastive term further improves final discrimination by encouraging old and new latent states to remain separated in the adapter space.
    }
    \label{fig:ablation_components}
\end{figure}

\paragraph{Necessity of the stable hidden state space.}
The largest performance gap in Table~\ref{tab:ablation_components} is between No Stable $H$ and Stable $H$ + Replay.
Without a stable latent state space, the model has no access to old-class samples or old-class state distributions after a new task arrives.
The classifier can still learn the newest task.
However, old decision regions collapse rapidly.
In the full CIFAR-100 Inc10 protocol, the final accuracy is only 9.06\%.
Introducing the frozen pretrained $H$-space changes the learning problem.
Old classes are no longer represented only by mutable network weights.
They are also represented by explicit prototype-centered latent distributions.
During each later task, sampled old states provide anchors for the adapter and classifier.
Thus, new-class learning is optimized together with old-class preservation.
This raises LastAcc from 4.55\% to 31.65\% in Inc5, from 9.06\% to 36.94\% in Inc10, and from 16.96\% to 43.16\% in Inc20.
These gains show that the stable hidden state is not just an implementation choice.
It is the main mechanism that makes latent replay meaningful.

\paragraph{Role of supervised contrastive separation.}
Latent replay restores old-class supervision.
However, replay alone does not explicitly control the geometry between old and new classes.
When a new class is visually similar to an old class, its adapted feature $z$ may still move into an old region.
This can happen even when old latent states are replayed.
The supervised contrastive loss addresses this failure mode.
It pulls same-class states together and pushes different-class states apart in the adapter space $Z$.
Empirically, adding the contrastive term increases LastAcc from 36.94\% to 37.06\% in the full CIFAR-100 Inc10 protocol and keeps nearly the same final accuracy in Inc5 and Inc20.
The gain is smaller than the gain from stable $H$-space replay.
Still, it is conceptually important.
It turns latent replay from a memory-preservation mechanism into a class-separation mechanism.
This is consistent with the method design.
$H$ provides a stable coordinate system for old-class world states.
The contrastive objective shapes the adaptive decision geometry where old and new classes coexist.
On full CIFAR-100, the contrastive term produces a small positive gain in Inc10 LastAcc and AvgAcc, while remaining nearly neutral in Inc5 and Inc20.
This suggests that contrastive separation is a complementary stabilizer rather than the dominant source of improvement.
The dominant innovation is the stable latent world memory; the contrastive loss refines its class geometry.

\subsection{Interpretability and Latent World Model Analysis}

We further examine whether the learned hidden states behave like reusable class distributions.
This analysis is important because our method aims to learn a latent world model, not only a label mapping.
A label mapping records a decision boundary for the current training data.
Once old samples disappear, this boundary can be overwritten.
There is no explicit mechanism to recover old-class evidence.
In contrast, a latent world model should preserve possible old-class states in a stable coordinate system.
If this view is meaningful, sampled old states should continue to constrain the classifier after many new tasks.
Old tasks should also remain recognizable, even though their raw images are never revisited.

\begin{figure}[t]
    \centering
    \includegraphics[width=\linewidth]{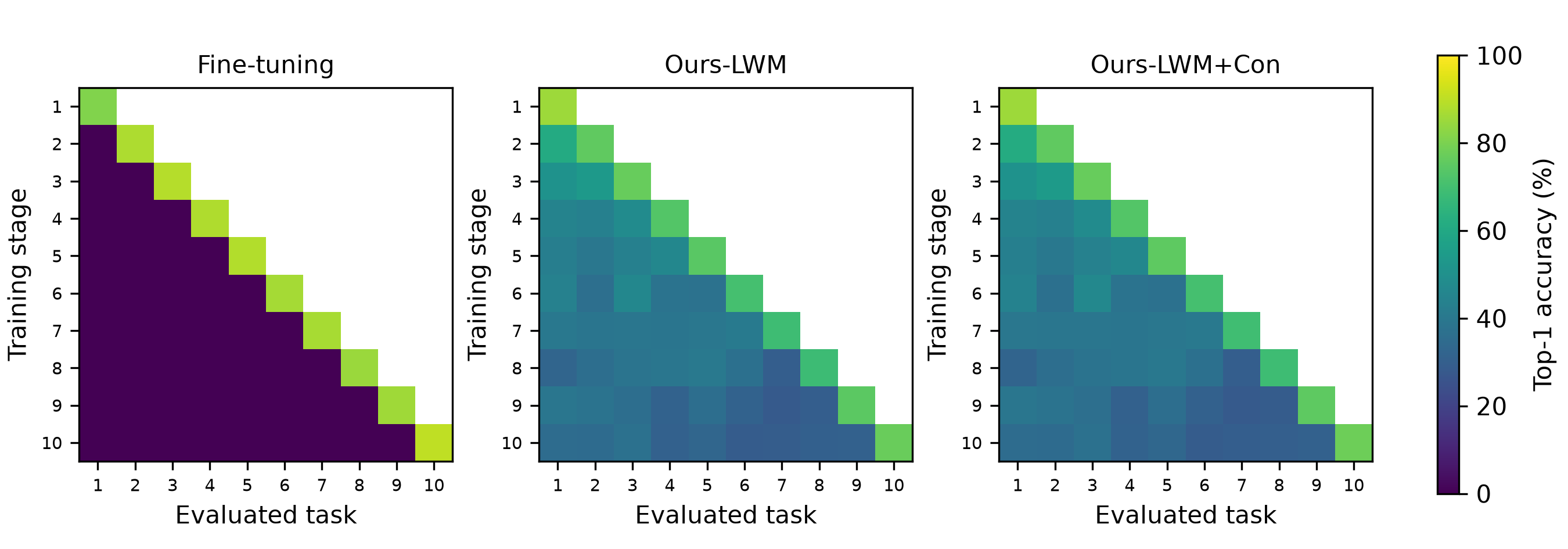}
    \caption{
    Task-retention heatmaps on full Split CIFAR-100 Inc10.
    Rows are incremental training stages and columns are evaluated tasks.
    Fine-tuning learns the newest task but old task columns rapidly collapse to zero.
    In contrast, latent world memory keeps non-zero accuracy across old task columns because old classes remain represented by replayable prototype distributions in the stable state space $H$.
    }
    \label{fig:latent_retention_heatmap}
\end{figure}

\paragraph{Task-retention matrix as a behavioral probe.}
We first use the per-task accuracy matrix as a behavioral probe.
It tests whether old-class latent distributions remain useful.
For Split CIFAR-100 Inc10, we evaluate the model separately on every seen task after each incremental stage.
The resulting matrix shows whether earlier class distributions are preserved or overwritten.
Figure~\ref{fig:latent_retention_heatmap} compares fine-tuning, Ours-LWM, and Ours-LWM+Con.

The heatmap shows a qualitative difference between the learning mechanisms.
Fine-tuning has high accuracy on the latest task.
At the final stage, however, its accuracy on previous tasks is nearly zero.
This is the signature of a current-task mapping.
The model can classify the classes it just saw, but it no longer has an operational representation of old classes.
Ours-LWM and Ours-LWM+Con instead retain visible mass across old task columns.
This means that old classes still occupy usable regions in the adapted decision space.
This remains true even though the model only receives sampled latent states from the stored world memory.

\paragraph{Old-state preservation at the final stage.}
Table~\ref{tab:latent_retention_summary} summarizes the final-stage behavior.
Fine-tuning obtains 90.60\% on the last task but 0.00\% mean accuracy on old tasks.
Its final overall accuracy is therefore only 9.06\%.
This confirms that high new-task plasticity alone does not imply continual recognition.
By contrast, Ours-LWM+Con preserves 32.52\% mean old-task accuracy while maintaining 77.90\% on the newest task.
The old-task accuracy is not perfect.
Still, it is clearly non-zero across many old tasks.
This is the behavior expected when old classes are represented as replayable latent distributions rather than unavailable raw images.

\begin{table}[t]
\centering
\caption{
Final-stage task retention on Split CIFAR-100 Inc10.
Old-task mean averages tasks 1--9 after all 10 tasks have been learned.
New-task accuracy is the final task accuracy.
}
\label{tab:latent_retention_summary}
\begin{tabular}{lccc}
\toprule
Method & Old-task Mean & New-task Accuracy & All-task Mean \\
\midrule
Fine-tuning & 0.00 & 90.60 & 9.06 \\
Ours-LWM & 32.46 & 77.30 & 36.94 \\
Ours-LWM+Con & 32.52 & 77.90 & 37.06 \\
\bottomrule
\end{tabular}
\end{table}

\begin{figure}[t]
    \centering
    \includegraphics[width=0.78\linewidth]{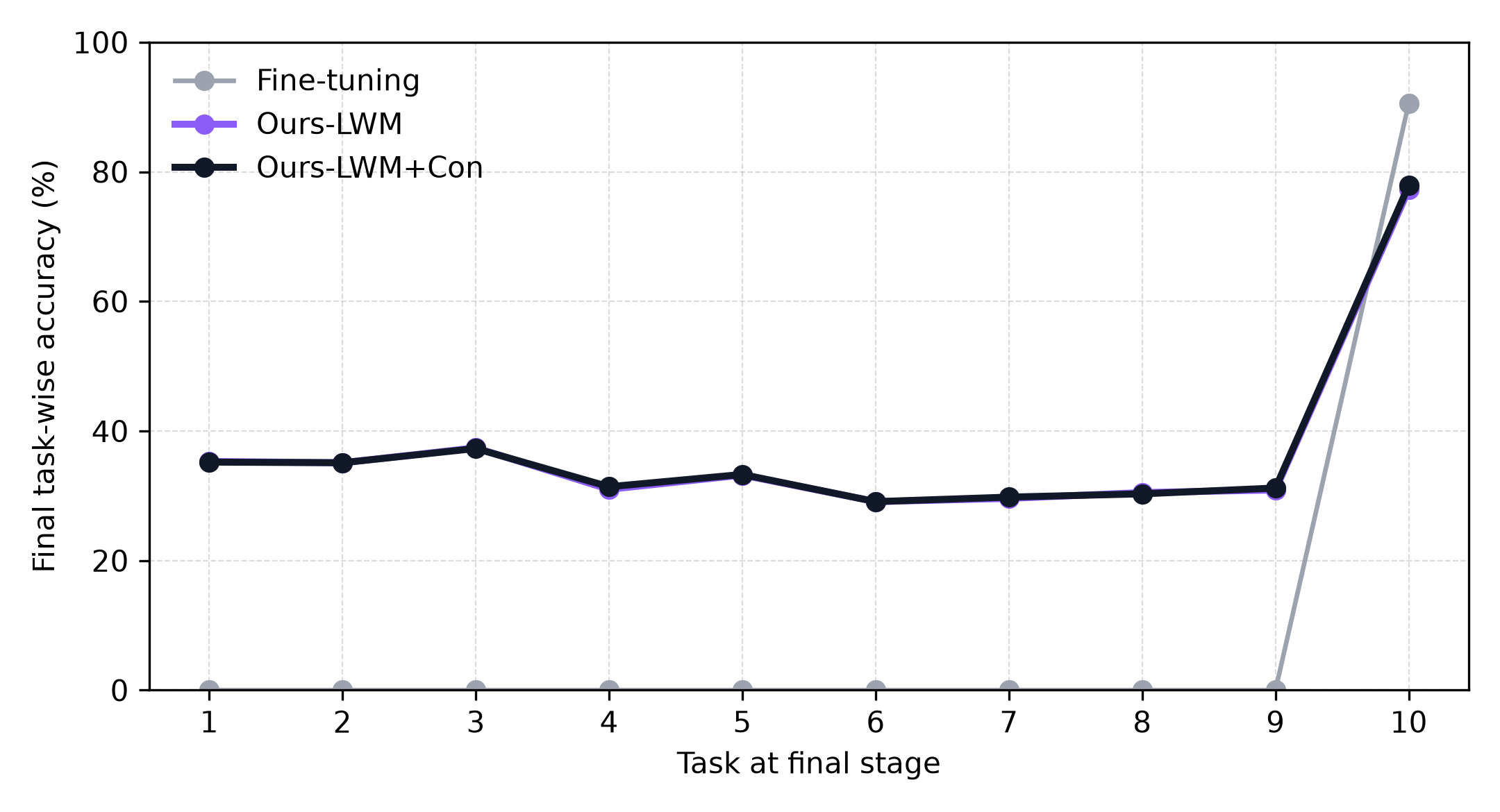}
    \caption{
    Final per-task accuracy profile on Split CIFAR-100 Inc10.
    Fine-tuning concentrates almost all accuracy on the newest task.
    Latent world memory distributes accuracy across both old and new tasks, indicating that old class distributions remain queryable after incremental learning.
    }
    \label{fig:latent_final_retention}
\end{figure}

\begin{figure}[t]
    \centering
    \includegraphics[width=\linewidth]{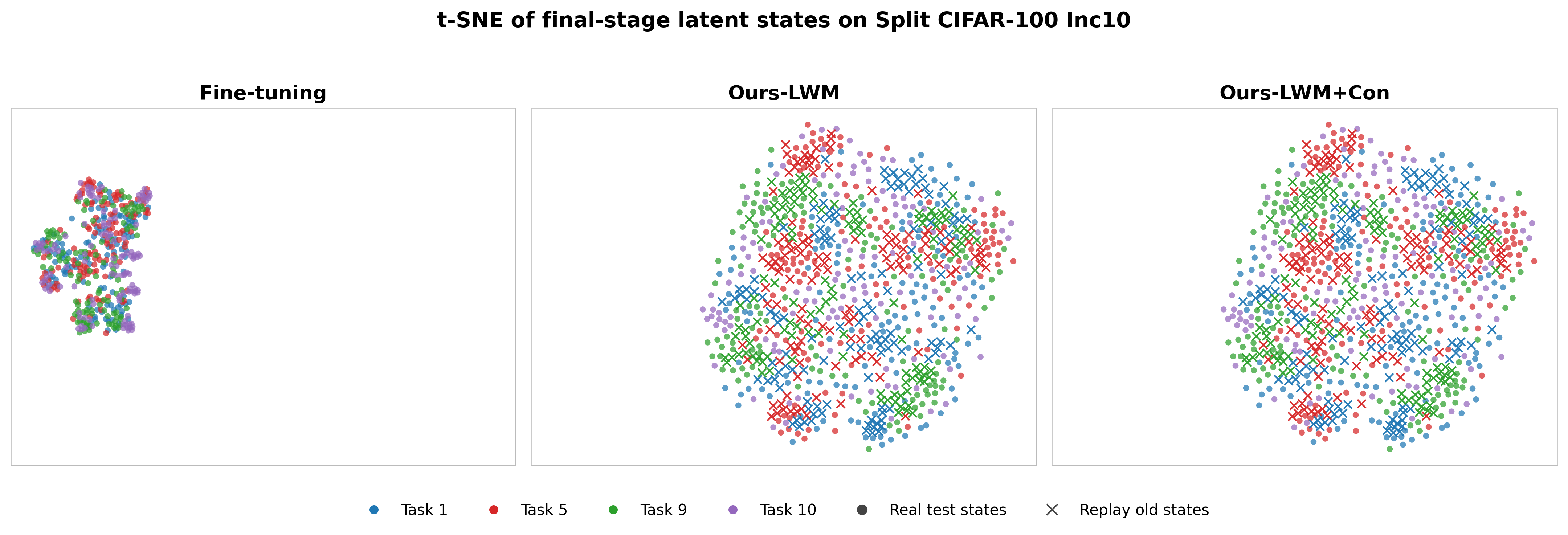}
    \caption{
    t-SNE visualization of final-stage latent states on Split CIFAR-100 Inc10.
    Colors correspond to four selected incremental tasks.
    Circles are real test hidden states and crosses are replayed old hidden states sampled from the prototype latent world memory.
    Fine-tuning has no replay distribution, while Ours-LWM and Ours-LWM+Con preserve replayable old-class states that overlap with real old-task regions.
    This indicates that the stable hidden space stores reusable class-state distributions rather than only a transient classifier mapping.
    }
    \label{fig:latent_tsne_comparison}
\end{figure}

\paragraph{t-SNE visualization of latent-state distributions.}
We further visualize the final-stage latent states.
The goal is to inspect whether the stored prototype world memory corresponds to meaningful old-class regions.
Figure~\ref{fig:latent_tsne_comparison} shows t-SNE embeddings of selected tasks from the final Split CIFAR-100 Inc10 model.
Circles denote real test samples and crosses denote old-class latent states sampled from the prototype memory.
Fine-tuning has no replayable old states; it only exposes the final adapted representations of real samples.
In contrast, Ours-LWM and Ours-LWM+Con generate old latent states that lie in the same broad regions as real old-task samples.
This visual evidence supports the distributional interpretation of the method.
The memory is not a label table.
It is a class-conditional latent state model that can still sample old-class evidence after later tasks have been learned.

\paragraph{Why this supports the distributional interpretation.}
The prototype memory does more than save one vector per label.
Each class is represented by a set of prototype-centered latent distributions in $H$.
Replay therefore samples multiple possible states around class modes.
This distributional view matters for three reasons.
First, it provides \emph{support}: old classes continue to produce training signals even when raw old images are absent.
Second, it provides \emph{geometry}: sampled states occupy old-class regions in the same stable coordinate system as new real features.
The adapter therefore learns old-new boundaries, not only new-class boundaries.
Third, it provides \emph{regularity}: because the encoder is frozen, the meaning of a stored prototype does not drift when future tasks are learned.

The results in Figures~\ref{fig:latent_retention_heatmap},~\ref{fig:latent_final_retention}, and~\ref{fig:latent_tsne_comparison} are consistent with this interpretation.
If the method only learned a transient mapping from current features to labels, old task columns would behave like fine-tuning.
They would collapse after later stages.
Instead, the old columns remain active.
This indicates that the stored latent distributions carry class-level state information.
The classifier can reuse this information later.
In this sense, the ``world model'' is not a generative image model.
It is a compact model of class states in the hidden world $H$.

\paragraph{Effect of class-separable contrastive geometry.}
The contrastive term has a more subtle role.
The retention matrix shows that the main source of old-state preservation is latent replay from stable $H$.
The contrastive variant slightly improves the final old-task mean from 32.46\% to 32.52\%.
It also improves final new-task accuracy from 77.30\% to 77.90\%.
This suggests that the contrastive loss mainly refines the boundary between old sampled states and new real states.
It does not replace the world memory.
Instead, it makes the replayed distributions more discriminative after they pass through the trainable adapter.
This supports our choice to separate the system into a stable state space $H$ for memory and an adaptive space $Z$ for class separation.

\section{Discussion}

The proposed method suggests that catastrophic forgetting can be reduced by preserving class distributions in a stable latent state space.
The object to be remembered in CIL does not have to be a raw image, a classifier weight, or a generated pixel-level sample.
Old knowledge can instead be represented as a distribution over hidden states.
This changes the interpretation of memory.
In conventional rehearsal, memory is a small subset of past observations.
In our framework, memory is a compact model of where a class lives in the semantic state space $H$.
The experimental results support this view.
With fine-tuning, old task accuracy collapses even though the newest task remains accurate.
With latent-state replay, old task regions remain queryable after many increments.

\paragraph{Meaning of the latent state space.}
The stable hidden state $H$ acts as a coordinate system in which classes can be described statistically.
This differs from treating features as transient intermediate representations.
Because the encoder is frozen, a stored prototype still refers to the same semantic region after later tasks are learned.
Thus, the prototype memory is not just a cache of vectors.
It is a compact approximation to the class-conditional state distribution.
This distinction is important for continual learning.
If the representation drifts, old distributions become invalid.
Latent replay may then harm learning.
If the representation is stable, old states can be sampled and reused as training evidence.
They give the adapter and classifier a persistent reference for old-class decision regions.
The separation between stable $H$ and adaptive $Z$ is therefore a main design principle.
$H$ stores the world, while $Z$ adapts the decision boundary.

\paragraph{Connection to self-supervised learning.}
Our current implementation uses supervised contrastive separation as the main auxiliary geometric constraint.
The latent-world perspective is also compatible with richer self-supervised objectives.
Methods such as BYOL, DINO, MAE, I-JEPA, and DINOv2 show that useful semantic states can emerge without class labels.
Future versions could train the encoder or adapter with self-supervised prediction, masked reconstruction, view consistency, or joint embedding prediction.
These objectives may make the hidden state distribution more robust before class labels are introduced.
For example, a JEPA-style objective could predict the latent state of one view or image region from another.
This would encourage $H$ to encode object-level regularities rather than task-specific label shortcuts.
It may be useful when new classes have few labels or when the stream contains unlabeled data between supervised increments.
Related cross-domain self-supervised and contrastive studies suggest that causal structure, disentangled contrastive transfer, and latent relation modeling can improve robustness under distribution shift \citep{li2025selfsupervisedfundus,liu2024graphdisentangled,ren2021lrgcn}.
Prior-knowledge-guided generation in 3D representation learning also suggests that structured world priors can organize latent semantic states beyond direct label supervision \citep{nie2024t2td}.

\paragraph{Toward zero-shot and unsupervised class discovery.}
A further extension is to use the latent world model for zero-shot or unsupervised class discovery.
If hidden states form meaningful class-conditional distributions, a new sample can be analyzed by how well it fits existing latent distributions.
One possible criterion is entropy.
When a sample is close to one known class distribution, its assignment distribution over prototypes or classes should have low entropy.
When it lies between old classes or belongs to an unseen category, the assignment distribution should become flatter and have higher entropy.
This provides a bridge between supervised CIL and the zero-shot self-supervised idea discussed above.
Category information may be inferred not only from labels, but also from the uncertainty structure of the latent state space.
In practice, one could combine prototype likelihood, nearest-prototype distance, and class-assignment entropy.
These signals could decide whether an incoming sample should be absorbed into an existing class, treated as a new class, or used for self-supervised adaptation.
Such an entropy-based mechanism could make the latent world model more autonomous.
It would not only preserve old classes, but also detect when the world has changed.
This direction is related to zero-shot and cross-modal retrieval, where disentangled representations transfer semantic structure to unseen categories \citep{li2026cdgan}.

\paragraph{Limitations and future work.}
The current approach depends on the quality of the pretrained encoder and the expressiveness of the prototype mixture memory.
If the frozen encoder does not separate semantically similar classes, the latent distributions may overlap.
Replay alone may then fail to prevent confusion.
Our memory model also uses relatively simple prototype-centered Gaussian sampling.
Future work may replace it with richer density estimators, normalizing flows, latent-space diffusion, or non-parametric uncertainty models.
Another direction is task-adaptive stability.
Rather than freezing all of $H$, the model could update part of the representation under constraints that preserve old latent distributions.
Finally, self-supervised entropy-based discovery could be integrated with supervised incremental updates.
This may move the system beyond fixed class-incremental benchmarks toward open-world continual learning, where labels and class boundaries are partly unknown.

\section{Conclusion}

We presented Prototype Latent World Model Replay, a class-incremental learning framework that treats old knowledge as distributions over stable hidden states.
It does not store raw exemplars or rely only on mutable classifier weights.
The core idea is to separate the system into two spaces.
The frozen pretrained latent state space $H$ provides a stable semantic coordinate system.
The adaptive classifier space $Z$ learns new classes while being constrained by replayed old states.
By storing each old class as a multi-prototype latent distribution, the model can sample old-class states during later tasks.
This helps preserve old decision regions without retaining raw old images.
The supervised contrastive constraint further shapes the adapter space so that old replayed states and new real states remain class-separable.

Experiments on Split CIFAR-100 show that the method reduces catastrophic forgetting across Inc5, Inc10, and Inc20 protocols.
The ablation study shows that stable latent-state replay is the main factor behind the improvement.
The contrastive loss adds a geometric regularizer for old-new separation.
The interpretability analysis further shows that retained hidden-state distributions remain behaviorally meaningful.
Old task accuracies remain non-zero after many increments, while fine-tuning collapses on old tasks.
These results support the view that a stable latent world model can serve as a compact memory for continual recognition.

More broadly, this work suggests a path from exemplar-based continual learning toward state-distribution-based continual learning.
Future systems may enrich the latent world model with stronger self-supervised objectives, entropy-based unknown-class discovery, and more expressive density estimators.
Such systems could not only remember old classes, but also detect and organize new semantic structure as the data stream evolves.

\bibliographystyle{plainnat}
\bibliography{references}

\end{document}